%% file: main.tex
\PassOptionsToPackage{hidelinks}{hyperref}
\documentclass[final]{clv2025}

\jvol{}
\jnum{}
\jyear{}

\dochead{Survey} 

\pageonefooter{}

\makeatletter
\newcommand{\applabel}[1]{%
  \addtocounter{section}{-1}%
  \refstepcounter{section}%
  \phantomsection%
  \label{#1}%
}
\makeatother

\makeatletter
\renewcommand\appendix{%
   \setcounter{section}{0}
   \renewcommand{\theequation}{\Alph{section}.\arabic{equation}}
   \renewcommand{\thefigure}{\Alph{section}.\arabic{figure}}
   \renewcommand{\thetable}{\Alph{section}.\arabic{table}}
   \renewcommand{\thesection}{\Alph{section}}
}
\makeatother

\usepackage{amsmath}
\usepackage{booktabs}
\usepackage{xcolor}
\usepackage{comment}
\usepackage{graphicx}
\usepackage[export]{adjustbox}
\usepackage{array}
\usepackage{varwidth} %
\usepackage{multirow}
\usepackage{tabularx}
\usepackage{cleveref}
\usepackage{makecell}
\usepackage{geometry}
\usepackage{todonotes}
\usepackage{subcaption}

\usepackage[acronym]{glossaries}
\makenoidxglossaries
\input{glossary}

\runningtitle{KD4MT}
\runningauthor{De Gibert, Attieh, Mickus, Scherrer, Tiedemann}

\renewcommand{\glossarysection}[2][]{}

\begin{document}

\title{KD4MT: A Survey of Knowledge Distillation for Machine Translation}

\author{
Ona de Gibert\thanks{Equal contribution. Author order determined via a coin toss.}$^{1}$, %
Joseph Attieh$^{* 1}$, 
Timothee Mickus$^{1}$,
Yves Scherrer$^{2}$,
Jörg Tiedemann$^{1}$
}

\affilblock{
 \affil{University of Helsinki\\\quad \email{firstname.lastname@helsinki.fi}}
 \affil{University of Oslo\\\quad \email{firstname.lastname@ifi.uio.no}}
}

\maketitle

\begin{abstract}

Knowledge Distillation (KD) as a research area has gained a lot of traction in recent years as a compression tool to address challenges related to ever-larger models in NLP.
Remarkably, Machine Translation (MT) offers a much more nuanced take on this narrative: in MT, KD also functions as a general-purpose knowledge transfer mechanism that shapes supervision and translation quality as well as efficiency.

This survey synthesizes KD for MT (KD4MT) across 105 papers (through October 1, 2025). 
We begin by introducing both MT and KD for non-experts, followed by an overview of the standard KD approaches relevant to MT applications.   
Subsequently, we categorize advances in the KD4MT literature based on (i) their methodological contributions and (ii) their practical applications.
Our qualitative and quantitative analyses identify common trends in the field and highlight key research gaps as well as the absence of unified evaluation practice for KD methods in MT. 
We further provide practical guidelines for selecting a KD method in concrete settings and highlight potential risks associated with the application of KD to MT such as increased hallucination and bias amplification. 
Finally, we discuss the role of LLMs in re-shaping the KD4MT field.
To support further research, we complement our survey with a publicly available database summarizing the main characteristics of the surveyed KD methods and a glossary of key terms.

\end{abstract}

\section{Introduction}
\label{sec:intro}

The performances of Large Language Models (LLMs) have been shown to follow scaling laws: increased size improves scores. 
This has propelled researchers in NLP to develop and publish models with ever larger parameter counts, with little regards for their environmental impact or computational requirements.
This trend has also been observed in the field of Machine Translation (MT) ever since the widespread adoption of neural MT models, and it has become more prominent recently with the release of large-scale LLM-based MT systems.%

Knowledge distillation (KD) is typically presented as a means to curb this race towards larger and computation-hungrier models: it is a compression technique that reduces 
the computational demands of large models while maintaining their performance. For this, KD draws inspiration from scholastic teaching paradigms: a powerful model, the \emph{teacher}, imparts its knowledge to a smaller model, the \emph{student}. However, KD has set itself apart from other compression techniques such as pruning and quantization \citep{park2024comprehensive}, as it can also be used to adapt models to specific tasks and domains. Such adaptations are especially interesting in the field of machine translation, where language and domain coverage can vary widely. In this survey, we provide a broad view on KD approaches and applications in order to demonstrate the versatile use of distillation methods in this particular sub-field of natural language processing, covering the extensive literature that has been accumulated over the past years.

\begin{figure}[h!]
 \centering
 \includegraphics[max width=0.5\linewidth]{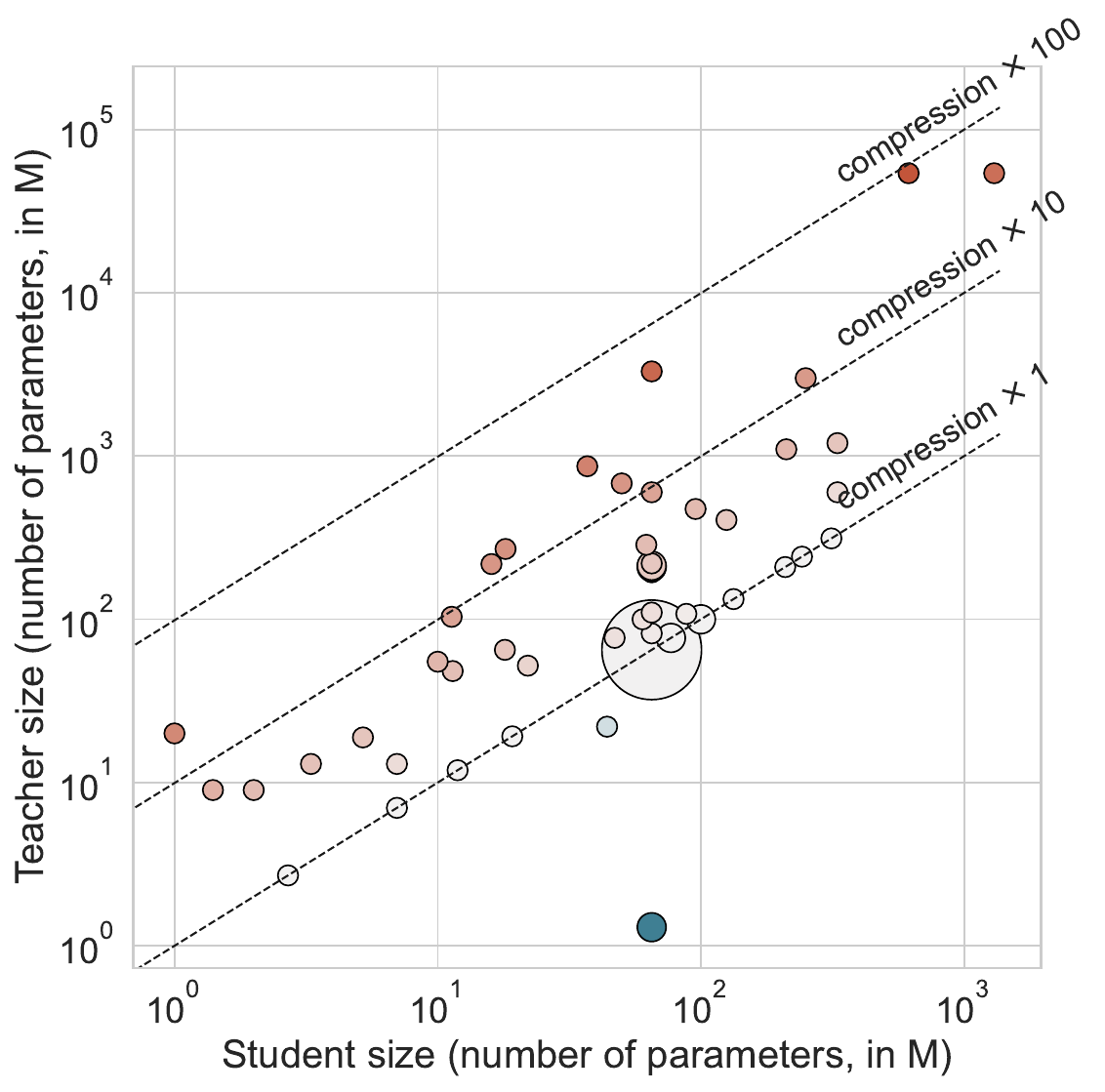}
 \caption{Teacher versus student parameter counts. Dot size indicates frequency of a specific configuration, color marks the compression ratio $\frac{\mathrm{Teacher\ size}}{\mathrm{Student\ size}}$. Roughly half of the works use a ratio of 1. Best viewed in color.
 }
 \label{fig:ratios}
\end{figure}

In total, we reviewed 105 papers presenting research on various aspects of knowledge distillation in the field of MT.\footnote{We considered papers published before October 1st, 2025.} \Cref{fig:ratios} plots the parameter counts of teacher and student models in those studies, demonstrating the variety of settings applied in the literature. Although we can point to a handful of works with students 50 times smaller than their teachers or more, we also observe that in almost half of the works we survey, teachers and students have equal numbers of parameters (with 28 out of 105 papers using a “transformer base” architecture for both teacher and student). This strongly suggests
that, at least insofar as MT as an application domain is considered, portraying KD as a compression technique first and foremost is inaccurate: in a large proportion of cases, KD is \textit{not} applied to bolster efficiency.

This calls for a broader view on KD in the light of MT. As we will see in the survey, many distillation techniques are applied for specific cases of knowledge transfer. For example, KD can be used to merge information from several models in order to increase language coverage, or on the contrary to specialize a general-purpose model to specific language pairs and domains. KD may be used to compensate for the lack of data and also to enable particular decoding strategies. KD is %
crucial for certain time-sensitive applications and acts as a regularizer and mode reduction technique. Translation-task-specific knowledge can be distilled from general-purpose language models and expand existing supervision signals.

To enable such broad application areas, many specific algorithms and methods need to be developed, for instance, by refining supervision or leveraging external sources for complementary information. 
We will shed more light on those specific approaches to KD in the field of MT. For readability, we divide the presentation into three parts: 
\begin{enumerate}
    \item A presentation of preliminaries related to machine translation and knowledge distillation (\Cref{sec:prelims})
    \item A discussion of specific KD algorithms for MT (\Cref{sec:algorithms})
    \item A summary of applications of KD in MT (\Cref{sec:apps})
\end{enumerate}

We conclude the survey with a discussion on the roles of KD considered in the literature, the limitations and gaps in the current literature and the increasing shift to LLM-based approaches also in the field of KD for MT in \Cref{sec:discussion}.

Our intent behind the present survey is two-fold. 
On the one hand, we wish to provide a convenient entry-point for readers knowledgeable in either MT or KD and wanting to familiarize themselves with another field.
In the interest of making the present article as thorough as possible, we also provide a glossary of key terms (in Appendix \ref{sec:glossary}) and a comprehensive database summarizing all methods and highlighting their specific characteristics (such as model sizes, compression ratios, datasets and evaluation metrics used, respective advantages and disadvantages).\footnote{Database available at \url{https://github.com/Helsinki-NLP/KD4MT-survey}.} 
On the other hand, we aim to underscore the concrete ways in which KD provides practical solutions for many problems underpinning MT applications, and conversely the advances in KD that are brought about by considering the characteristics of MT as an NLP task.

\section{Preliminaries}
\label{sec:prelims}
We begin with a high-level overview of MT and KD. We refer the reader to \citet{koehn-2020-neural} and \citet{gou2021knowledge} for in-depth introductions.

\subsection{The Task of Machine Translation}
\label{sec:intro_mt}
Machine Translation (MT) is the task of generating target language output that conveys the same information as some given source language input.\footnote{Note that we restrict the task to text-based translation in this survey.} In Neural Machine Translation (NMT), we model translation as a conditional probability $P(\mathbf{y}|\mathbf{x})$ over target language tokens $\mathbf{y}=(y_1,y_2,...,y_T)$ given a sequence of source language tokens $\mathbf{x}=(x_1,x_2,...,x_S)$, %
where the conditional probability is factorized as %

\begin{equation}
    p_{\theta}(\mathbf{y}\mid \mathbf{x})
\;=\;
\prod_{t=1}^{T} p_{\theta}(y_t \mid \mathbf{y}_{<t}, \mathbf{x})
\end{equation}
where $\mathbf{y}_{<t} = (y_1, y_2, \dots, y_{t-1})$ refers to the context before time step $t$. This property, i.e., conditioning each token on all previously generated tokens $\mathbf{y}_{<t}$, is known as \textit{autoregressive}.

In training, the objective is to learn the model parameters ($\theta$) that maximize the probability of target sentences ($\mathbf{y}$) (or other units of language) given their corresponding source sentences ($\mathbf{x}$), across a corpus of parallel translation pairs $D=\{(\mathbf{x^{(i)}}, \mathbf{y^{(i)}})\}_{i=1}^{N}$. This is done by minimizing the Negative Log-Likelihood (NLL), which is equivalent to maximizing the likelihood:

\begin{equation}
    \mathcal{L}_{\mathrm{MLE}}(\theta)
\;=\;
-\!\!\sum_{(\mathbf{x},\mathbf{y})\in D}
\log p_{\theta}(\mathbf{y}\mid \mathbf{x})
\end{equation}

This is typically implemented using the Cross-Entropy Loss ($\mathcal{H}$) accumulated over all time steps

\begin{equation}
\mathcal{L}_{\mathrm{CE}}(\theta)
\;=\;
\sum_{(\mathbf{x},\mathbf{y})\in D}
\sum_{t=1}^{T}
\mathcal{H}\Big(\delta_{y_t},\, p_{\theta}(y_{t} \mid \mathbf{y}_{<t}, \mathbf{x})\Big)
\end{equation}
where $\delta_{y_t}$ is the ground-truth distribution at $t$ (a one-hot vector with probability 1 at the correct token $y_t$) and $p_{\theta}$ is the predicted probability distribution over the vocabulary at time step $t$ (the position in the target sequence) given by the model ($p_{\theta}$). During training, at each time step, the model is fed the ground-truth previous token as input, rather than its own predictions. This approach allows efficient parallelization without sequential dependencies, which also enables a stable gradient flow without error accumulations from previous predictions.

The traditional architecture for NMT is the encoder-decoder transformer \cite{vaswani2017attention}, a sequence-to-sequence model that decouples the mapping of input text to intermediate context representations %
(done by the encoder) from the generation of the output text from these representations (decoder). Both parts heavily rely on multi-head self-attention in the transformer architecture, while cross-lingual attention connects the decoder layers with the contextualized embeddings of the final encoder layer. In the decoder, future tokens are masked in training to learn the auto-regressive nature of the model, generating target text from left to right.
At inference, decoding in autoregressive MT is typically performed with beam search, which expands the output sequence step by step by keeping only the most probable tokens.

\paragraph{Non-autoregressive translation (NAT)} 
Autoregressive models require that decoding is strictly sequential and has to proceed one token at a time. In contrast, non-autoregressive translation approximates the search with conditionally independent token generation steps

\begin{equation}
   p_{\theta}(\mathbf{y}\mid \mathbf{x})
\;\approx\;
\prod_{t=1}^{\hat T(\mathbf{x})} p_{\theta}\!\left(y_t \mid \mathbf{x}\right)
\label{eq:nat}
\end{equation}
where the sequence length $\hat T(\mathbf{x}) = P_L(x)$ is predicted by a separate length prediction module $P_L$. This enables parallel generation of tokens, thus making NAT suitable for latency-critical environments, but at the expense of ignoring target-token dependencies.
An overview of NAT generation methods in NMT can be found in \citet{xiao_etal_2023_natsurvey}.
As we will see, KD is crucial for making NAT work, and we review the corresponding works in \Cref{sec:kd for nat}.

\paragraph{LLM-based translation}
In recent years, multilingual large language models (LLMs) have increasingly been used for MT. Most often, LLMs rely on a decoder-only architecture, 
in which the condition on the source language input is given by the prompt and the translation is generated as a continuation of the input string. Source and target language are concatenated and typically augmented by instructions using templates such as the following from \citet{luukkonen-etal-2025-poro}

\begin{quote}
    \texttt{<|user|>}Translate into Finnish: \{srctxt\} \texttt{<|assistant|>} \{trgtxt\}
\end{quote}

\noindent
where the input text and translated texts are inserted into {\em \{srctxt\}} and {\em \{trgtxt\}}.
As is common in LLM training, pre-training is done on large corpora using simple next-word prediction objectives over the entire sequence, and subsequent supervised fine-tuning (SFT) is typically added to improve the model's instruction-following capabilities. One of the biggest differences between task-specific MT models %
and LLM-based MT is the emerging ability of LLMs to translate arbitrary text units such as paragraphs or entire documents without being explicitly trained for such context sizes. %
Their increasing translation performance makes them the preferred tool also for cross-lingual tasks, but at the expense of raising inference costs. As we will see in Section \ref{sec:discussion_llms}, most surveyed papers focus on KD for task-specific MT models, but the use of KD in LLM-based translation is becoming increasingly frequent. Such applications will be covered in \Cref{sec:enriching}.

\subsection{Foundations of KD for MT}
\label{sec:background}

KD is a compression technique first introduced by \citet{bucila2006model} and later coined by \citet{hinton2015distilling}. It has been historically referred to as the process of transferring the knowledge from a powerful teacher to a compact student, by minimizing the divergence between teacher and student output distributions. KD is distinct from other model compression techniques in that it requires two training processes: (i) training the teacher model, and (ii) training the student model with supervision of the trained teacher model. In contrast, other compression techniques such as pruning or quantization are typically decoupled from the training process and applied to a fully trained model. 

From a methodological perspective, KD approaches can be broadly classified into two categories: response-based methods and feature-based methods \citep{gou2021knowledge}. \textbf{\gls{response-kd}} refers to the transfer of knowledge from the final prediction of the teacher to the student. In \textbf{\gls{feature-kd}}, the knowledge is transferred from the intermediate layers of the teacher.

\begin{figure*}[t!]

  \centering
  \begin{subfigure}[t]{0.32\textwidth}
    \centering
    \includegraphics[width=\linewidth,trim=80pt 80pt 380pt 20pt,clip]{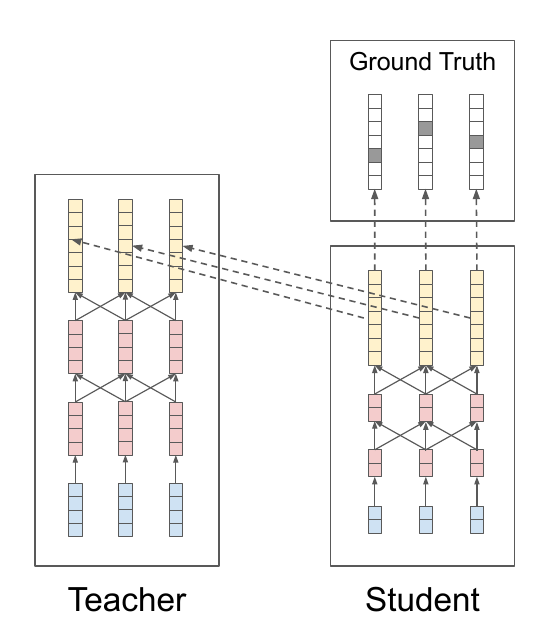}
    \caption{Word-level KD}
  \end{subfigure}\hfill
  \begin{subfigure}[t]{0.32\textwidth}
    \centering
    \includegraphics[width=\linewidth,trim=80pt 80pt 380pt 20pt,clip]{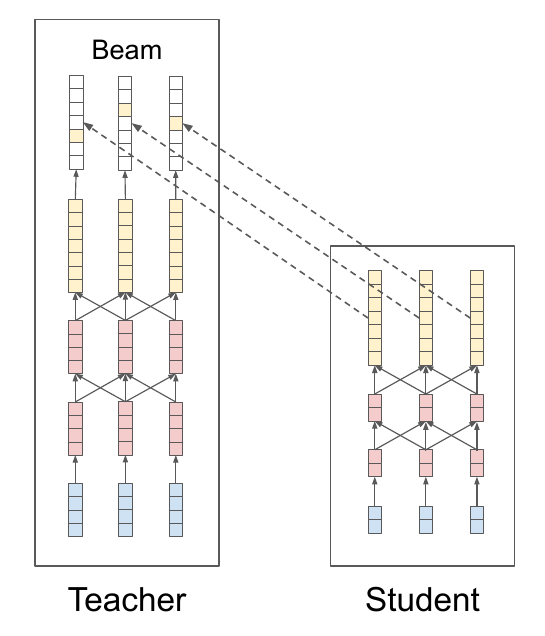}
    \caption{Sequence-level KD}
  \end{subfigure}\hfill
  \begin{subfigure}[t]{0.32\textwidth}
    \centering
    \includegraphics[width=\linewidth,trim=80pt 80pt 380pt 20pt,clip]{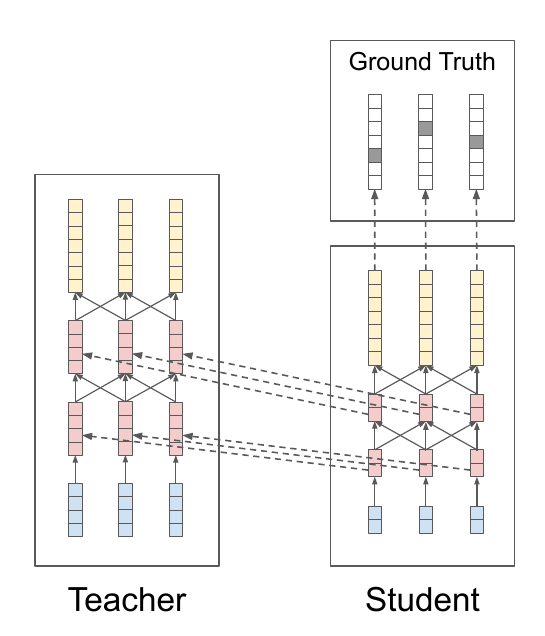}
    \caption{Feature-based KD}
  \end{subfigure}
   \caption{Overview of the main KD methods for MT \citep[adapted from][]{kim-rush-2016-sequence}. The blue rectangles symbolize input embedding vectors, the red rectangles intermediate representations of (typically) transformer layers, and the yellow rectangles the probability distributions over output tokens. The dashed lines represent the supervision signal that is provided by the different KD approaches: the teacher provides the student with output distributions (word-level KD), decoded sequences (sequence-level KD), or intermediate representations (feature-based KD). Best viewed in color.} 
 \label{fig:kd_methods}
\end{figure*}

KD was first considered in the context of MT by \citet{kim-rush-2016-sequence}. KD for MT sets itself apart from other NLP tasks, since it must generate full sequences, with each prediction conditioned on the previously produced tokens. This makes response-based methods more attractive for MT applications: they focus on the final predicted translations of the teacher and are thus more straightforward to implement. Consequently, \citet{kim-rush-2016-sequence} propose two main response-based approaches for MT, \textbf{\gls{word-kd}} and \textbf{\gls{seq-kd}}, whereas 
feature-based methods have only appeared more recently in the context of MT. An overview of the three main methods of KD for MT is shown in \Cref{fig:kd_methods}.%

Below, we introduce the core KD methods that form the foundation for the method variants discussed in \Cref{sec:algorithms} and the applications explored in \Cref{sec:apps}.
KD encompasses many named variants; for clarity, we define them in the Glossary in \Cref{sec:glossary}, 
including terms not covered in the main text.%

\subsubsection{Response-based KD}
\label{sec:response-based-kd} 

Following \citet{kim-rush-2016-sequence}, the two main response-based approaches for MT are \glslink{word-kd}{Word-Level} and \gls{seq-kd}. As the names suggest, they differ by the level of granularity they operate at.

\paragraph{Word-Level KD (Word-KD)} The student model is trained to output a similar distribution as the teacher model on every token. 
Assuming access to a teacher probability distribution, \citet{kim-rush-2016-sequence} minimize the cross-entropy between the output distributions of the student and the teacher at the token level. %
Equivalently, this objective can be written as minimizing the Kullback--Leibler (KL) divergence from the teacher to the student. It should be noted that KL is an asymmetric function, as it penalizes discrepancies differently depending on which distribution is treated as the reference, thereby influencing how the student model aligns with the teacher’s output distribution.  The loss for Word-KD can be formulated as:

\begin{equation}
\mathcal{L}_{\mathrm{KD}}(\theta_s;\theta_t)
\;=\;
\sum_{(\mathbf{x},\mathbf{y})\in D}
\sum_{t=1}^{T}
D_{\mathrm{KL}}\!\Big(
p_{\theta_t}(y_{t} \mid \mathbf{y}_{<t}, \mathbf{x})
\;\big\|\;
p_{\theta_s}(y_{t} \mid \mathbf{y}_{<t}, \mathbf{x})
\Big)
\end{equation}
\noindent where $\theta_t$ and $\theta_s$ denote the parameters of the teacher and the student, respectively.

In practice, the student receives two types of supervision at each training step: the output probability distribution of the teacher, and the reference token given by the training data. The training objective for the student involves a convex combination of this loss with the standard cross-entropy loss from the observed data, weighted by a parameter $\alpha$. This parameter typically lies within the range [0, 1] and is selected through hyperparameter optimization.
The final loss is then formulated as:

\begin{equation}
\mathcal{L}_{\text{Word-KD}}(\theta_s;\theta_t)
\;=\;
(1-\alpha)\,\mathcal{L}_{\mathrm{MLE}}(\theta_s)
\;+\;
\alpha\,\mathcal{L}_{\mathrm{KD}}(\theta_s;\theta_t)
\label{eq:word-kd}
\end{equation}

Word-KD is also referred to as \textbf{soft distillation}, as it trains on the soft probability distribution over the vocabulary; or \textbf{online distillation}, as the teacher’s targets are produced during training (on-the-fly).

\paragraph{Sequence-Level KD (Seq-KD)} 
The student model is trained to mimic the behavior of the teacher at the sentence level. In MT, sequence-level training typically involves minimizing the Negative Log Likelihood (NLL), where the model aims to match the predicted tokens with the one-hot distribution of the reference tokens, evaluating how well the predicted probability distribution aligns with the true target distribution. Since this loss with respect to the teacher involves a summation over an exponentially large number of possible sequences, \citet{kim-rush-2016-sequence} simplify the problem by 
using beam search to generate the best candidate sequences on the training set. The loss for Seq-KD can be formulated as:
\begin{equation}
\mathcal{L}_{\text{Seq-KD}}(\theta_s;\theta_t)
\;=\;
-\!\!\sum_{\mathbf{x}}\,
\log p_{\theta_s}\!\left(\hat{\mathbf{y}} \mid \mathbf{x}\right)
\end{equation}  
\noindent where $\hat{\mathbf{y}}$ is the output from running beam search with the teacher model.

In practice, Seq-KD is reduced to a two-step process: first, the teacher model generates target sequences by translating the source training data, and second, the student model is trained on this synthetic dataset. This means that Seq-KD does not require the student training data to be parallel. Hence, Seq-KD can be considered a \textbf{data augmentation} approach, since it relies on artificially generated training data from the teacher model. %

Seq-KD can also be referred to as \textbf{hard distillation}, as it trains on hard one-hot labels; or \textbf{offline distillation}, as it is a two-step procedure and the teacher targets can be produced beforehand.

 \subsubsection{Feature-based KD}
\label{sec:feature-based-kd}
As mentioned previously, feature-based KD transfers knowledge from the intermediate representations of the teacher model to the student model. This is done by matching either hidden states, feature or attention maps of the teacher with the student's via any loss function designed to measure representational similarity, such as cosine similarity or mean-squared error.   
The loss for feature-based KD can be formulated as:
\begin{equation}
\mathcal{L}_{\text{Feat-KD}}(\theta_s;\theta_t)
\;=\;
(1-\alpha)\,\mathcal{L}_{\mathrm{MLE}}(\theta_s)+ \alpha\,\mathcal{L}_{F}\big( f_{\theta_t}(\mathbf{x}),\, f_{\theta_s}(\mathbf{x}) \big)
\end{equation}
\noindent where $f_{\theta_t}(x)$ and  $f_{\theta_s}(x)$ are the intermediate representations of teacher and student models, respectively, and $\mathcal{L}_F$ is the similarity function used.  In practice, the student can also receive supervision using the cross-entropy loss from the observed data, weighted by a parameter $\alpha$.

Feature-based KD offers an alternative to the more commonplace response-based KD algorithms. As such, it does provide more flexibility in terms of how and where knowledge from the teacher is distilled into the student model. On the other hand, it also comes with its own sets of caveats; in particular, intuitions developed for response-based systems are not always applicable to feature-based algorithms.

\section{Algorithms of KD for MT}
\label{sec:algorithms}

KD has become a central strategy for either compressing or transferring knowledge in MT. In \Cref{sec:background}, we presented the main KD methods and showed how they refer to different ways of transferring knowledge from the teacher to the student. But what this general presentation does not touch upon is the amount and type of knowledge that is being transferred. Here, we provide an overview of different variants of KD algorithms that adapt the knowledge transfer process to specific needs of MT.

Moving beyond the canonical \textbf{one-teacher one-student one-language pair} setting, recent work shows that KD succeeds or fails largely by differentiating \emph{what we distill from}.
We frame KD algorithms for MT as the process of engineering the supervision signal across three dimensions: (i) selecting supervision by controlling what to distill, (ii) expanding supervision through multi-teacher, proxy-task, or LLM-based approaches, and (iii) reframing supervision beyond classical training paradigms. \Cref{tab:section3} lists the main research questions tied to these three dimensions. We address them in turn in the following subsections and summarize our main findings in \Cref{sec:algo_summary}.

\begin{table*}[h]
\centering
\small
\begin{tabular}{p{3.4cm} p{4.5cm} p{3.8cm}}
\toprule
\textbf{Algorithmic Strategy} & \textbf{Research Question} & \textbf{Solution Strategy} \\
\midrule
Selecting Supervision \newline (\Cref{sec:algo_filtering}) & What knowledge should be transferred, and at what granularity? & Knowledge Filtering \\
\midrule
Expanding Supervision \newline (\Cref{sec:enriching}) & How can knowledge be enriched beyond single-teacher supervision? & Multi-Teacher Distillation \newline Proxy-Task Distillation \newline Distillation from LLMs \\
\midrule
Reframing Supervision \newline (\Cref{sec:assymetric}) & How should knowledge be aligned under distributional and behavioral mismatches between teacher and student? & On-Policy Distillation \newline Symmetric Distillation \\
\bottomrule
\end{tabular}
 
\caption{Algorithmic variants of KD for MT, organized by the core research question each class of methods addresses.}
\label{tab:section3}
\end{table*}

\subsection{Selecting Supervision}%
\label{sec:algo_filtering}

The effectiveness of KD in MT can benefit from selectively controlling \emph{what} teacher knowledge should be transferred to the student.  Rather than applying distillation indiscriminately, multiple works have proven that better student models are trained when inaccurate and uncertain information is filtered out from the supervision during KD. This selection can operate at various levels of granularity: at the token level for Word-KD, at the sentence level for Seq-KD and at representation level for feature-based KD.

 \paragraph{At the token level (Word-KD methods)} KD has increasingly shifted from treating all tokens uniformly to removing supervision signals whenever they are not needed. In fact, traditional Word-KD applies a KL divergence loss between the teacher and student distributions across every token, ignoring how learning difficulty varies across the sequence. Recent research aims instead to estimate the token-level difficulty and adjust the distillation factor accordingly.  Some estimate it using the discrepancy between model predictions and reference translations \cite{wang-etal-2021-selective, song-etal-2025-self}, while others estimate it through model calibration \cite{lee2022hard}. Alternatively, \textbf{Top-1 Information Enhanced Knowledge Distillation} (TIE-KD; \citealp{zhang-etal-2023-towards-understanding}) simplifies the process altogether, arguing that most of the teacher's useful knowledge lies in its most confident predictions, and thus only transfers the probability mass from the top-1 token at each position. Complementary to such \emph{filtering}, \textbf{\gls{annealing-kd}} \citep{jafari-etal-2021-annealing} progressively lowers the temperature of the teacher’s output softmax distribution during training.  Initially higher temperatures provide softer, easier-to-learn distributions, and gradually shifting towards sharper distributions aligns training complexity with the student’s improving capability.  However, as a stand-alone strategy, annealing distillation (a form of soft curriculum learning) is outperformed by most methods in the literature earlier \cite{zhang-etal-2023-towards-understanding,xu-etal-2025-self-distillation}, and is therefore not recommended on its own.

 \paragraph{At the sentence level (Seq-KD methods)} Traditional Seq-KD trains students directly on teacher-generated translations. However, relying on all samples from the teacher-generated distillation set is often suboptimal, as the teacher predictions are not always correct. Consequently, recent research has focused on refining the distillation set, by exploring ways to select or synthesize examples that maintain faithfulness to the reference.  Early work used \gls{seq-interp} \cite{kim-rush-2016-sequence}, generating multiple translations per source sentence and keeping the one best aligned with the reference using sentence-smoothed BLEU \citep{chen2014systematic}. Later studies \cite{freitag2017ensemble,zhang-etal-2018-analyzing}  directly evaluated each teacher output, removing poor candidates or replacing them with their corresponding gold references, using Translation Edit Rate \cite{snover-etal-2006-study} or embedding similarity to the source sentences \cite{zhang-etal-2018-analyzing}. Building on this, more recent work \cite{zouhar-2021-sampling} showed that relying on a single teacher output is suboptimal, and that combining multiple hypotheses, upsampling, and mixing synthetic with original data produces stronger students than using only one hypothesis per source. In fact, at the sentence level, generating multiple candidates per source sentence and then filtering is more effective than relying on a single output, as it provides diversity and retains more candidate translations. Single-output filtering risks over-filtering and losing valuable knowledge; therefore, replacing poor candidates with their corresponding gold references can help preserve translation quality.

 \paragraph{At the representation level (feature-based methods)} Knowledge filtering determines which parts of the internal representation of the teacher should be transferred to the student.  Different approaches have explored this selection: some target entire layers \cite{wu-etal-2020-skip}, others focus specifically on the attention heads \citep{jin2024align}, and yet others operate on the whole parameter space of the model \cite{lin-etal-2021-weight}. Building on these ideas, \textbf{iterative adjacent-layer matching} \cite{xu-etal-2025-self-distillation} devised a \glslink{self-kd}{self-distillation} method that distills knowledge from each teacher layer into the preceding student layer, effectively turning the teacher into a sequence of student layers. 
Selecting knowledge on the representation level is much less common, more complex to implement and less reliable. For instance, layer-wise supervision \citep{wu-etal-2020-skip} may be more effective for teacher and student models with similar architectures,  while the method proposed by \citet{jin2024align} may be better suited for models with different architectures. At the same time, iterative adjacent-layer matching has shown to be more effective than the SOTA in response-based KD (TIE-KD; \citealp{zhang-etal-2023-towards-understanding}). %

\subsection{Expanding Supervision}%
\label{sec:enriching}
While standard KD relies on a single teacher to supervise a single student, recent research has explored how to enrich or diversify the supervision signal available to the student. In this context, \emph{enriching} refers to broadening and deepening the supervisory signal that the student model receives during KD, going beyond the limited information contained in a single teacher’s output. %
These strategies share a common goal to make the student’s learning process  richer. %

\paragraph{Multi-Teacher Distillation}
Recent research shows that exposing a student to multiple teachers enriches the supervision signal and boosts downstream performance \cite{9722996}, as it diversifies the signal transmitted to the student.
Using multiple teachers can be done by averaging their probability distributions using Word-KD \cite{freitag2017ensemble} or aggregating the respective distilled datasets using Seq-KD \cite{li-etal-2019-niutrans}. More instances of such work are found in \Cref{sec:apps}. To avoid training multiple teachers for the task, this effect can be modeled with a single teacher by introducing  structural and architectural diversity to the KD process: \gls{mutual-distillation} \cite{miao-etal-2023-exploring} extracts several students of different sizes from the teacher and trains them in parallel, allowing the best-performing student at each step to act as a temporary co-teacher and provide additional gradients to its peers.  Likewise, \citet{9722996} achieve similar diversity by dynamically altering the teacher’s architecture during its training (through layer reordering, pruning, or dropout) to create multiple subnetworks that serve as implicit teachers during distillation, thus avoiding the cost of training separate models. %

\paragraph{Proxy and Auxiliary Task Distillation}
Proxy (intermediate) tasks improve the distillation by providing learning signals that go beyond the teacher’s raw logits, typically through an auxiliary objective such as masked language modeling (MLM) or its conditional variant (CMLM).  Instead of relying solely on translation supervision, these methods integrate pretrained or fine-tuned language models into the MT pipeline to infuse richer linguistic knowledge and functional diversity to the supervision signal. For instance, BERT-based approaches fuse bidirectional lexical representations into the encoder–decoder framework, enabling the MT model to benefit from contextualized token-level semantics \cite{Weng_Yu_Huang_Cheng_Luo_2020, chen-etal-2020-distilling, yoon2023emnetwork}. Others extend this idea by coordinating training between MT and language modeling components, such as concerted training (CTMT) \cite{yang2022making}, which balances pretrained knowledge and task adaptation through dynamic gating and rate scheduling. Beyond MLM-based proxies, Teacher Assistant enhanced KD (TAeKD; \citealp{lv-etal-2024-taekd}) broadens supervision by aggregating outputs from multiple close-sourced translation systems using a proxy fusion model to approximate soft targets.%

\paragraph{LLMs}
Large language models (LLMs) have been increasingly utilized in KD4MT, either to generate distilled datasets (under black-box Seq-KD, refer to \Cref{sec:lrmt}) or to supervise the knowledge distillation process without participating in constructing the distilled dataset. 
Using LLMs enriches the supervision signal by (i) diversifying candidate translations and enabling Minimum Bayes Risk (MBR)-style implicit re-ranking \cite{finkelstein2024mbrqefinetuningtrainingtime,wang2024dontthrowawaydata}, (ii) diagnosing and correcting systematic errors by synthesizing contrastive samples (source, faulty translation, corrected translation) \cite{li2024mtpatcher}, and (iii) providing rationale-level guidance that can be distilled alongside translations \cite{wu-etal-2025-boosting}.  Using LLMs elevates KD from a static teacher–student framework to a knowledge-rich interaction, where LLMs not only transfer linguistic competence but also encode reasoning and quality awareness, ultimately driving more generalizable and interpretable MT systems. However, despite their promise, these approaches face notable limitations. Heavy reliance on LLMs introduces computational and cost overheads, as high-quality supervision often requires large prompts, multiple inferences, or MBR sampling.

\subsection{Reframing Supervision}%
\label{sec:assymetric}

Performing KD inherits two major issues. First, standard KD training paradigms introduce a distribution mismatch between the teacher and the student, known as the \gls{capacity-gap}.%
This occurs when the size mismatch between the teacher and the student, usually a smaller network, leads to the student being unable to learn how to properly fit the data. 
This phenomenon can be observed when the student either spreads probability mass over all dimensions (mode-averaging) or collapses onto a few teacher peaks (mode-collapse).
Relatedly, the \gls{capacity-curse} can also take place, where a larger network does not always yield a better student.
Second, the autoregressive training approach commonly used in MT leads to a training–test discrepancy, often referred to as exposure bias due to teacher forcing,\footnote{During training, the decoder is conditioned on the ground-truth previous token, whereas at inference, it must rely on its own predicted token.} which causes errors to accumulate progressively during sequence generation.

To address these issues, some KD methods shifted away from the classic supervised training paradigms and borrowed concepts directly from reinforcement learning (RL): 
\textbf{f-distill} \cite{wen-etal-2023-f} replaces asymmetric KL loss with RL-style symmetric rewards that penalize both over-smooth and over-peaked distributions, encouraging the student to match the teacher’s full probability distribution and eliminating the distribution mismatch.  \textbf{Imitation-KD (ImitKD)} \cite{lin-etal-2020-autoregressive} borrows from imitation learning to train the student on its own sampled prefixes, with the teacher providing guidance on those specific contexts, thus reducing exposure bias. 
Extending both approaches, \textbf{Generalized KD (GKD)} \cite{agarwal2024onpolicy} adopts a fully on-policy regime, \footnote{ On-policy learning is an approach in reinforcement learning in which the data used to train the policy is generated by the same policy currently being improved.} where the student learns from its own predictions while leveraging the teacher’s soft targets through flexible divergence measures. GKD can also plug in any
divergence measure from \citet{wen-etal-2023-f}, inheriting the benefits of f-distill. Empirical evidence shows that such on-policy training strategies yield more stable and generalizable MT models, underscoring that the closer the training process mirrors inference conditions, the more effectively knowledge is distilled.

\subsection{Takeaways from KD Algorithms}\label{sec:algo_summary}

We have seen three main ways the literature modifies the canonical setup of KD. First,  some works focus on \emph{filtering} the learning signal, by keeping the most useful tokens or sentences. Second, other works try to \emph{diversify} supervision by adding external sources of knowledge, with multiple teachers, auxiliary tasks or LLM-based feedback.
Finally, more recent work moves beyond standard KD methods and \emph{reframes} supervision by borrowing ideas from RL to train on the model’s own predictions and to use symmetric divergences.

Together, all these lines of work aim to refine supervision, but from different perspectives.
The first two sets of methods aim to improving knowledge transfer, while the latter is focused on correcting structural shortcomings of KD, such as exposure bias and capacity mismatches.
RL-based methods are more expensive to train, so they should be used only when clear gains are expected.
It should be noted that cost concerns affect all KD methods in general, yet are almost never addressed.
Word-KD and Seq-KD are usually preferred across the first two groups of methods, while Seq-KD is typically applied for reframing supervision. However, few works use feature-based KD, even though it currently achieves SOTA results \cite{jin2024align}.
Therefore, it is crucial to empirically evaluate the methods to determine the most effective approach, keeping in mind the model architectures and task requirements.
Lastly, we remark that while the methods we have surveyed in this section tackle different ways to strengthen the training signal, they are in practice complementary and compatible with one another. Nonetheless, they are hardly ever used together.

\section{Applications of KD in MT}
\label{sec:apps}
Up until now we have reviewed the algorithmic variants of the standard KD methods. 
Another way to look at KD is by considering its specific use-cases in MT-related settings. We have identified four main application areas within MT where KD has been used: 
Multilingual MT, Low-Resource MT, Domain Adaptation, and Time-Sensitive Settings. %
These applications all have their unique challenges, as illustrated in \Cref{tab:section4}, and therefore have pushed KD in specific directions to meet them.

\begin{table*}[h]
\centering
\small
\begin{tabular}{p{2.2cm}p{4.5cm}p{5.3cm}}
\toprule
\textbf{Application} & \textbf{Research Question} & \textbf{Solution Strategy} \\
\midrule
Multilingual MT \newline (\Cref{sec:multilingualmt}) & How can KD mitigate the curse of multilinguality? & Multiple teachers \newline Multilingual teachers \\
\cmidrule{2-3}
& How can KD help deploy massively multilingual MT models under resource constraints? & Scale up and distill \\
\midrule
Low-resource MT \newline (\Cref{sec:lrmt}) & Where does the learning signal come from when strong teachers and parallel data are missing? & Monolingual data as soft teacher \newline Pivot through high-resource language \newline Pre-trained LMs as teachers \\
\cmidrule{2-3}
& How can KD help deploy MT models under computing resource constraints? & KD for compression \\
\midrule
Domain adaptation \newline (\Cref{sec:domain_adap}) & How can KD mitigate catastrophic forgetting and domain shift? & Split and distill \newline Multi-domain KD \\
\midrule
Time-sensitive settings \newline (\Cref{sec:time_apps}) & How can we inject future target knowledge for non-standard decoding strategies? & Future-aware MT to compensate for partial input in simultaneous MT \newline 
Seq-KD-simplified data for NAT to enable low-latency MT \\
\bottomrule
\end{tabular}
 \caption{Applications of KD in MT organized by the core research questions.}
 \label{tab:section4}
\end{table*}

\subsection{Multilingual MT} %
\label{sec:multilingualmt}
Multilingual MT (MMT) refers to the translation task where multiple languages are involved either at the source or target side of a translation model. 
This is typically done by training a single model on the concatenation of training data covering all language directions \citep{johnson2017google}. 
For a detailed overview on the topic of MMT, we refer the reader to \citet{dabre2020survey}.

A major issue that arises in MMT is the so-called \emph{\glslink{multilinguality-curse}{curse of multilinguality}}, i.e., the tendency of a single multilingual MT model to lose per-language quality as the number of languages increases, because fixed model capacity is spread thinner across the languages \cite{conneau2020unsupervised}. Empirically, the curse appears when a multilingual model underperforms separate bilingual systems trained for the same language pairs. %
To mitigate the curse of multilinguality, one option is to train strong bilingual teachers and distill them into a single multilingual student  \citep{tan2019multilingual}. %
Because related languages benefit  from transfer learning \citep{dabre-etal-2017-empirical}, an alternative is to distill from multilingual teachers. This transfers knowledge more effectively than many separate bilingual teachers %
and can yield consistent gains independently of whether the languages belong to the same language family or not, surpassing distillation from bilingual teachers \cite{doo-dee-2023-target}.  %

An alternative way of circumventing the curse of multilinguality is to scale up the capacity of MMT models: Massively Multilingual MT (MMMT; \citealp{aharoni-etal-2019-massively}) models handle hundreds of languages and often require billions of parameters to achieve satisfactory translation quality across languages. 
Such large models are challenging to deploy, and KD has been proposed as an effective model compression technique. Up to now, there is no evidence in the literature favoring Seq-KD or Word-KD methods for MMMT compression. %
As for deciding on the architecture of the student, deep encoders with shallow decoders are commonly adopted due to better performance and efficiency at inference \cite{kasai2020deep, bapna2022building, mohammadshahi2022small100, galaindictrans2}.

\subsection{Low-resource MT }
 \label{sec:lrmt}
Low-resource MT (LRMT) refers to the task of translation that involves languages with limited amounts of training data. 
LRMT can also be framed as a translation task in a limited-budget setting, where computational resources and internet access are scarce.
In the first case, one of the most common approaches when dealing with low-resource languages (LRLs) is to use multilingual models \citep{arivazhagan2019mnmt}. In the latter, compression techniques are needed to produce compact, efficient models suitable for deployment under constrained environments.

In the context of KD for data-scarce LRMT, the traditional teacher-student paradigm is not always applicable simply because it is not possible to obtain a sufficiently strong teacher model. Hence, we notice that the research focus lies less on the specific KD variants used, but rather on answering the following question: Where does the learning signal come from when strong teachers and parallel data are missing?

Whenever a strong teacher is unavailable, the learning signal to train the student has to come from what data is available; whether the data is monolingual, pivoted or synthetic.
 When parallel data and a conventional teacher are absent, a model trained on monolingual data only can act as a soft teacher for the low-resource target language. These approaches exploit structure (distributional similarity, fluency) rather than supervised alignments, turning LRMT into a form of indirect target modeling. This is done by building a target distribution to distill from using lexical similarity on target vocabularies \cite{zhang2020improving}, using LM-regularized distillation objectives where a target-side LM shapes student distributions \cite{baziotis-etal-2020-language}, or distilling contextual encoders (e.g., BERT) into the student \cite{zhang2024distilling}. 

Another approach is to use a high-resourced pivot language to supply the missing signal by training and chaining two models trained respectively on source–pivot and pivot–target corpora. However, naive two-hop cascades amplify error propagation from the first hop to the second \cite{cheng2017pivot}. KD changes this dynamic by using pivoting to distill training targets, transforming  the cascade into a single-step supervision for the student and thereby reducing propagated errors \cite{chen2017teacher,ahmed2024neural}. \citet{Yang_2022} go further by training a single multilingual teacher on all available pivot–source and pivot–target corpora. %
One-step supervision from the multilingual teacher aggregates information across pairs (transfer learning) while avoiding the compounding errors of two-hop generation.  Complementarily, \citet{he2019language} organize supervision as a graph where languages are nodes and MT models are edges; low-resource pairs absorb signal from higher-accuracy neighbors via forward and backward Seq-KD, turning related languages into structured pathways for transfer.

Additionally, it has become increasingly common to use pre-trained language models, whether for synthetic data generation or Word-KD supervision: rule-based MT (RBMT) offers low-variance deterministic anchors to train on \cite{chang2025integrating,de-gibert-etal-2024-hybrid} while LLMs, though weaker than dedicated MT teachers, can act as synthetic data generators and provide parallel data where none exists \cite{oh2023data,yang2023neural,de2025scaling}.
While fully synthetic data (both at the source and target sides) has proven beneficial \cite{aji2021fullysyntheticdataimproves}, relatively few works examine how to generate it most effectively. Overall, increasing the diversity on the decoding side, e.g., in terms of sampling functions \citep{galiano-jimenez-etal-2025-beyond} or by corrupting the target \citep{galiano2025fluent}, generates students with higher translation quality. 
Furthermore, post-distillation finetuning of pre-trained encoder-decoder teachers (mBART, M2M-100, NLLB) on LRLs can help transfer cross-lingual coverage into compact students \cite{galiano2023exploiting,de-gibert-etal-2023-four}.  
However,  this commonly hurts the model performance on the high resource languages it was trained on, which can be mitigated through careful pruning and re-training \cite{zhang2023importance}. 
When multiple teachers are available, the strategy is to decide how to coordinate between the teachers instead of picking  only one.  Rather than relying on typology heuristics to pick the teacher (sometimes helpful, \citealp{dabre-etal-2017-empirical}; sometimes not, \citealp{kocmi-bojar-2018-trivial}), data-driven weighting learns who should teach whom, either by adaptively combining bilingual teachers during distillation \cite{saleh2020collective} or by first forming cluster-level assistants trained on different language groups and then distilling hierarchically \cite{saleh-etal-2021-multilingual-neural}.

On a last note, as mentioned above, LRMT can also refer to the translation task in a resource-constrained environment. Moreover, the availability of linguistic resources for a language often goes hand in hand with the availability of computational resources in the regions where this language is spoken. %
If MT models are to be deployed by the local communities, this raises the need of efficient and small models. %
This can be achieved using Seq-KD to train students at a strong quality–speed trade-off \cite{dabre-fujita-2020-combining}. Multiple interesting directions emerge: 
Seq-KD often yields higher accuracy than quantization but is more sensitive to data size, model choice, and teacher confidence, whereas quantization is more stable and consistent \cite{diddee-etal-2022-brittle}.
Furthermore, post-distillation fine-tuning on high-quality data reliably helps the student \cite{gibert-etal-2025, gumma2023empirical}, while narrower and deeper students can match wider and shallower ones with fewer parameters \cite{gumma2023empirical}.\footnote{Consistent with findings from \Cref{sec:multilingualmt}} %

\subsection{Domain Adaptation}
\label{sec:domain_adap}

The goal of domain adaptation is to modify a model trained on a source (out-of-domain) corpus in such a way that it performs well on a new target (in-domain) domain. The main challenge in domain adaptation is \emph{catastrophic forgetting}, i.e., the risk of overfitting the model to the target domain such that it ``forgets'' useful out-of-domain knowledge, limiting its robustness.
For a full overview of domain adaptation, we refer the reader to \citet{saunders2022domain}.

All KD-based approaches in this context separate general knowledge from domain-specific knowledge and design learning and transfer around that split, what we name the split-and-distill approach.
Regarding single domain adaptation, one line of work trains one model per domain and transfers between them, either bidirectionally and iteratively \citep{zeng-etal-2019-iterative,liu2022neural} or sequentially, distilling first from a general-domain teacher and then from an in-domain teacher into the same single student \citep{gordon-duh-2020-distill}.
Another approach structurally isolates the two sources inside a single model, by distinguishing between the parameter subsets that drive general versus in-domain performance through pruning
\citep{gu-etal-2021-pruning}.

In multi-domain adaptation, as the number of domains increases, we observe similar KD solutions as before, that rely on separating the knowledge per domain and distilling them into a single model.
This can be seen as a mirror of the approach by \citet{tan2019multilingual} for multilingual MT. Multi-domain adaptation can be performed using Word-KD by distilling every teacher solely on the data belonging to its domain \citep{mghabbar2020building}, and using Seq-KD \citep{currey-etal-2020-distilling} by synthesizing training data from each teacher on its domain data and training the student on the mixture. Another option is to isolate each domain in its own gated adapter and use KD to guide a classifier that selects which adapter to activate at inference \cite{klimaszewski2023gated}.\footnote{Dropping domain labels (i.e., clustering the data, fine-tuning one teacher per cluster, and distilling) showed that knowledge from diverse domain experts can be effectively combined using Seq-KD without explicit domain supervision \cite{currey-etal-2020-distilling}}

A line of work within domain adaptation studies continual learning with KD. The goal is to add knowledge from new domains while preserving what was learned from previous domains, even when the original domain data is no longer available.
\citet{liang-etal-2024-continual} leverage unlabeled source-side data with a contrastive objective aimed at  establishing inter- and intra-domain similarities in the model’s representations.%

\subsection{Time-Sensitive Applications}
 \label{sec:time_apps}
By time-sensitive applications, we refer to a group of MT paradigms that extend beyond conventional autoregressive translation and take time into account, namely: (a) simultaneous translation (SiMT; \citealp{miao-etal-2021-generative}), which is latency-sensitive and (b) non-autoregressive MT (NAT; \citealp{gu2018non}), which enables parallel decoding. What these approaches share is that they break or modify the standard left-to-right, full-sentence, autoregressive assumption. %

Together, these applications highlight how KD can effectively address unique MT-specific time-sensitive requirements. However, it is important to note that the role of KD is different across the two paradigms. While SiMT systems can be trained without Seq-KD (the works presented here simply leverage KD as a technique for improving performance), Seq-KD is required to enable NAT models.

\subsubsection{KD for SiMT}
In \Cref{sec:intro_mt}, we defined the task of MT and its left-to-right autoregressive objective. This objective enforces sequential decoding, where each token depends on the source and the past targets, but such a decoding strategy is  unable to account for right-to-left future target token dependencies. To overcome this, future-aware MT enhances translation by anticipating upcoming content, improving relevance and coherence.

Future-aware MT allows models to be exposed to future tokens at training time even though this is not possible at inference time. Future-aware student models use KD to add look-ahead or bidirectional target context to left-to-right MT, boosting coherence despite the absence of future tokens at test time. Concretely, they differ in (i) where future context comes from: future-aware LM \cite{zhang2019future,zhuang-tu-2023-pretrained,zhou-etal-2022-confidence} or bidirectional "seer" decoder with past-only and future-only sub-decoders \citep{feng-etal-2021-guiding}; (ii) what is transferred: hidden states \citep{zhang2019future} vs. token distributions \citep{feng-etal-2021-guiding,zhou-etal-2022-confidence,zhuang-tu-2023-pretrained}; and (iii) where it is transferred: to the decoder only \citep{zhang2019future,feng-etal-2021-guiding,zhou-etal-2022-confidence} vs. both encoder and decoder \citep{zhuang-tu-2023-pretrained}. %

SiMT is a practical application that benefits from the improvements brought along by future-aware supervision during training.
SiMT requires real-time translation with only partial source input, as seen in live interpretation, requiring monotonic and low-latency translations. The training of SiMT usually follows a \textit{wait-k} policy (\citealp{ma-etal-2019-stacl}), where a model is trained to translate using only the first $t-k$ source tokens when producing the $t$-th target token. SiMT systems are known to hallucinate more than full-sentence MT \citep{yu-etal-2025-investigating}.  
KD for SiMT pursues two main learning objectives: (i) monotone-compatible targets and (ii) future-aware internal representations.

On the data side, Seq-KD can be used to learn target sequences that are compatible with the SiMT task: distilled targets are produced by incremental wait-$k$ decoding  \cite{wang-etal-2023-better} or via SiMT-aware sampling that favors longer and more monotone predictions \cite{deng2022improving}. Beyond target construction, jointly training the SiMT student with a non-SiMT teacher injects ``future'' knowledge about upcoming source and target tokens while preserving monotonicity,  using Feature-based KD alone \cite{zhang2021future} or together with Word-KD \cite{yu-etal-2025-investigating}. Furthermore, self-training using KD can increase monotonicity and stabilize outputs as source tokens arrive by lowering output entropy \cite{sen-etal-2023-self}.

\subsubsection{KD for NAT}
\label{sec:kd for nat}
As introduced in \Cref{sec:intro_mt}, non-autoregressive MT (NAT) aims at accelerating translation by generating output sequences in parallel rather than sequentially. This is motivated by some drawbacks of regular autoregressive models \cite{gu2018non}: high inference latency due its sequential token generation, diminishing returns with increased beam size \cite{koehn-knowles-2017-six}, and computational dependency between beams.
The most commonly used NAT models are the Levenshtein Transformer \cite{gu2019levenshtein}, a NAT model that generates a sentence by editing it based on Levenshtein distance operations: insertion and deletion; and Mask-Predict \cite{ghazvininejad2019mask}, which generates a translation by iteratively masking and filling tokens. For a comprehensive overview of NAT, we refer the reader to \citet{xiao_etal_2023_natsurvey}.

NAT approaches rely on Seq-KD \cite{lee-etal-2018-deterministic}: typically, a source dataset is forward-translated using an autoregressive (AT) teacher and used to train a NAT student. 
NAT benefits from KD to maintain quality despite reduced latency.
Seq-KD is used such that the teacher generates more deterministic, simplified, single-reference translations that reduce target variability and are therefore easier to learn. We return to the regularization abilities of KD in Section ~\ref{sec:regularization}.

The most KD-relevant focus area is how to construct the distilled datasets. %
Prior work improves distillation by curating more effective training sets, for example, proposing methods such as generating multiple distilled references \cite{shao2022one}, mixing raw and distilled data \cite{guo2021self, liu2023selective}, or leveraging large amounts of monolingual data \cite{zhou2020improving}.
While curated strategies improve the MT performance, the resulting datasets often remain overly deterministic, which makes NAT struggle with low-frequency word prediction \cite{ding2021rejuvenating,ding2021understanding}. To overcome this, \citet{DING2025101765} propose an architecture-agnostic approach for NAT to better learn lexical choices without introducing any computational cost based on \gls{data-diversification} \cite{NEURIPS2020_7221e5c8}.

\subsection{Takeaways from KD Applications}
In this section, we have seen %
how KD can bring concrete solutions to problems inherent to specific use cases of MT. It can combat the curse of multilinguality by distilling stronger teachers into a single student; and it can address the catastrophic forgetting issue by balancing performance across domains.
It also compensates for the lack of resources in LRMT, by drawing additional supervision signal from existing external sources. Finally, it supports time-sensitive scenarios by transferring knowledge that enables efficient or constrained decoding without sacrificing too much quality.

Across applications, there is overlap both in the KD methods the community explores and in the types of solutions KD provides. %
With respect to methods, the use of Word-KD and Seq-KD is spread across applications, with a considerable amount of Seq-KD works in LRMT, where specific tooling makes it easy to deploy (e.g., Stopes, \citealp{andrews-etal-2022-stopes}; OpusDistillery, \citealp{opusdistillery}). By contrast, the study of feature-based methods remains very limited in these areas, with only a few instances in future-aware MT and SiMT. %
In addition, most works rely on standard KD methods, without applying further optimizations or variants such as those discussed in Section \ref{sec:algorithms}.

Regarding proposed solutions, works in both multilingual MT and LRMT often target compression explicitly.
Deep encoders and shallow decoders, paired with post-distillation fine-tuning, usually provide a good balance between quality and efficiency. %
Moreover, insights from low-resource and time-sensitive setups are consistent: increasing diversity in synthetic data generation improves downstream performance.
Lastly, multilingual MT and domain adaptation are closely related: multilingual MT can be viewed as a form of domain adaptation, where each language pair acts as a different domain. As a result, techniques developed for multilingual MT are often applied to domain adaptation (e.g. \citealt{mghabbar2020building} apply \citealt{tan2019multilingual}), and, conversely, domain adaptation methods are sometimes transferred to multilingual settings (e.g. \citealt{zhao2022life} apply \citealt{chuang-etal-2020-lifelong}).
These overlaps suggest that the community would benefit from a more unified view of KD applications in MT, such as the one presented in this survey.

\section{Discussion}
\label{sec:discussion}

 \label{sec:discussion_tendencies}
In this survey, we have approached the field of KD for MT from different angles: the main methods have been presented in \Cref{sec:prelims}, some key algorithmic improvements in \Cref{sec:algorithms}, and the most challenging applications in \Cref{sec:apps}. This may make it hard to see the big picture: Which methods are used for which applications? Which ideas have been most popular? Has the focus of KD4MT shifted over time?

\begin{figure*}[ht]
  \centering
    \centering
    \includegraphics[width=0.5\linewidth]{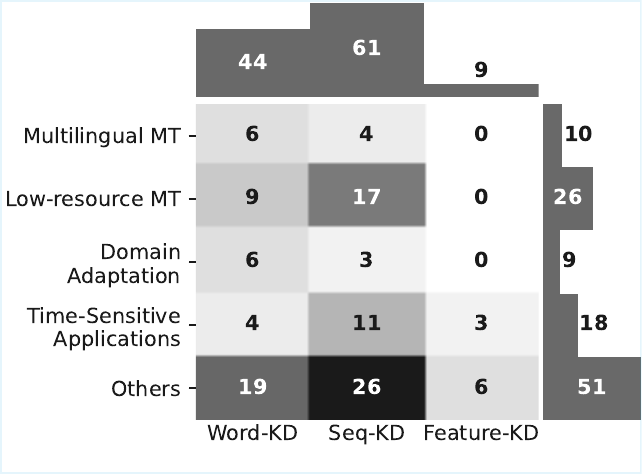}
    \caption{Breakdown of works surveyed per application and KD algorithm used. We surveyed 105 papers. However, some studies proposed methods utilizing more than one fundamental KD method, resulting in 115 unique approaches overall. }
    \label{fig:breakdown-heatmap}
\end{figure*}

In \Cref{fig:breakdown-heatmap}, we provide a broader overview of the current state of KD4MT, by classifying all surveyed works according to the applications they target and the methods they use. It becomes apparent that response-based KD, and in particular Seq-KD, is remarkably more popular than feature-based KD. Specifically, Seq-KD has established itself in two specific areas: as a data augmentation strategy for low-resource MT, and as a way to generate more deterministic and ``easier-to-learn'' data for non-autoregressive student models. A considerable amount of surveyed papers do not specifically target one of the four applications presented in \cref{sec:apps}. Typically, these papers focus on methodological innovations such as those identified in \Cref{sec:algorithms}, relying on widely used MT tasks (one language pair of relatively high-resourced languages) for their experiments. Some of the works in the \textit{Others} category are comparative studies, which are described in Sections~\ref{sec:regularization} and \ref{sec:discussion_roads}.

\begin{figure}[h]

    \centering
    \includegraphics[width=\linewidth]{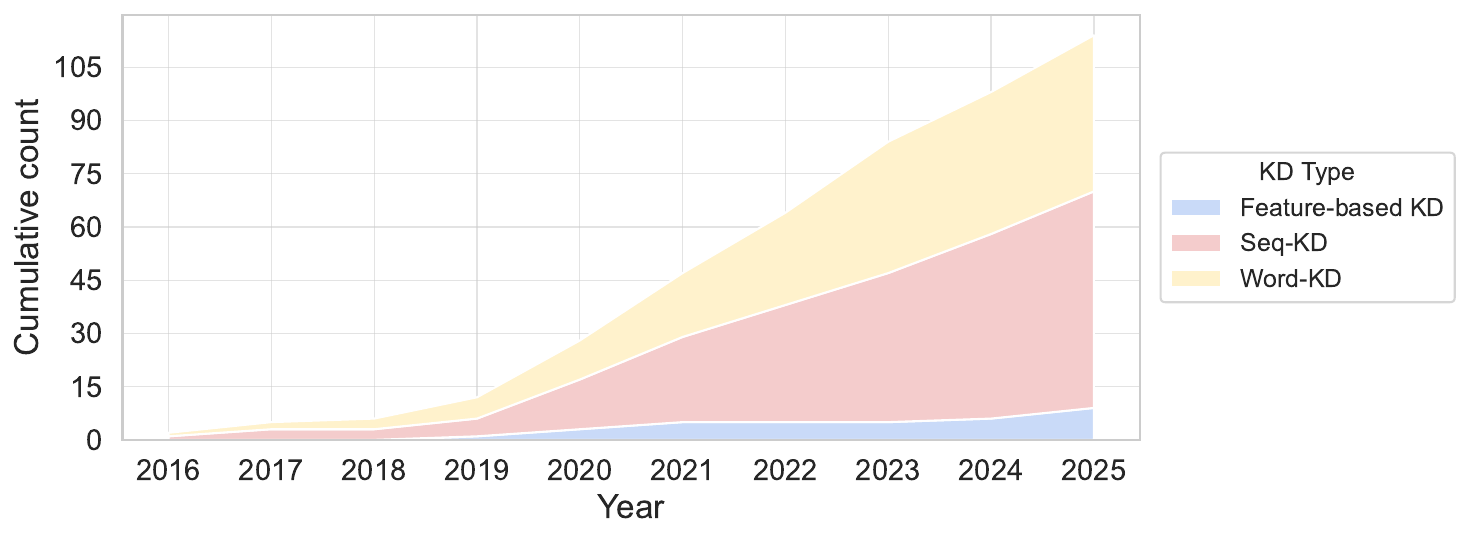}
    \caption{Cumulative number of papers using each KD type (Word-KD, Seq-KD, Feature-based KD) from 2016–2025.}
    \label{fig:timeline}
\end{figure}

\Cref{fig:timeline} provides the cumulative count of papers published since 2016 according to KD type. The number of publications in KD4MT started rising remarkably in 2019. At first, this rise was mostly driven by Seq-KD, with Word-KD approaches becoming more popular recently. As observed earlier, feature-based KD remains marginal in MT settings.

\subsection{Practical Considerations}
 \label{sec:discussion_pracitcal}

From the works reviewed above, we have seen how choosing the best KD method is not a trivial task. %
Nonetheless, some practical considerations can help guide the choice of a KD method: %
the access to teacher models, the available compute budget and the type of data all influence this choice.

A first consideration is whether one has access to the \textbf{model internals} of the teacher.  Word-KD requires access to the teacher's token-level probabilities; Feature-based KD requires access to its intermediate activations. Seq-KD is the most flexible option with respect to teacher access -- it only requires the teacher's output strings. This makes Seq-KD the only option that can distill from black-box commercial or API-based teachers. For example, there is a range of works that use Seq-KD for distilling from closed-source GPT-family models \cite{oh2023data,yang2023neural,de2025scaling}.

A second take considers the \textbf{architectural compatibility} between teacher and student. Word-KD  requires identical output vocabularies (and thus, tokenizers) between the two,\footnote{Practical workarounds that align distributions across different vocabularies exist \cite{boizard2025towards}. However, standard Word-KD formulations assume matching vocabularies.} while Feature-based KD needs a mapping between layers but not completely identical architectures. In contrast, Seq-KD imposes no architectural or vocabulary constraints beyond the student being trainable from the teacher's text outputs.

However, if one is working on a restricted \textbf{computational budget}, Seq-KD is generally more expensive on the teacher side. %
Word-KD only requires forward passes to obtain per-step distributions or features, whereas Seq-KD typically requires full decoding (e.g., beam search) to produce sequences, which is more computationally expensive. %
Furthermore, feature-based KD methods often incur a large memory footprint because their loss functions operate over logits from both the teacher and the student for each mini-batch.

The choice between Word-KD and Seq-KD often reflects a tradeoff between performance and computational cost. For instance, the \citet{costa2022nllb} find that Seq-KD performs marginally better in a small scale setup, but nevertheless choose to distill  their very large scale multilingual model (a 54B Mixture of Experts model) through Word-KD for reasons of computational efficiency. 
At the same time, the \textbf{reuse} of Seq-KD generated data is more straightforward, as illustrated by the emergence of new repositories targeted specifically at synthetic data, such as SynOpus.\footnote{\url{https://opus.nlpl.eu/synthetic/}}

Finally, depending on the availability of the type of training data, i.e., whether it is parallel or monolingual, supervision options differ regarding \textbf{ the use of ground truth}. Word-KD and Feature-based KD are often combined with ground-truth supervision, mixing cross-entropy with KD via the hyperparameter weight $\alpha$, as per \cref{eq:word-kd}. However, all of the KD methods can be applied without target-side ground truth labels, enabling training with monolingual or unlabeled data when references are unavailable. %

In sum, Seq-KD is generally the most flexible option. It can be used when teacher access is limited or architectures mismatch. However, if on a tight budget, one should prefer more efficient approaches such as Word-KD. %
If only monolingual data is available, one may use Seq-KD or feature-based KD.
However, recommendations about feature-based KD should be treated cautiously, since this area has received relatively little attention. 

Besides these general guidelines, if the goal of applying KD is compression, one should review the works in Sections \ref{sec:multilingualmt} and \ref{sec:lrmt}. 
Although KD is often framed as a compression technique, it is used in many more ways, as illustrated in \Cref{fig:ratios}. In our survey, only a few works target compression specifically. In many cases, KD serves other purposes. 

\subsection{KD for regularization rather than compression}
\label{sec:regularization}

A natural follow-up question is to consider what engineering goals KD \emph{does} fulfill.
In the context of NAT, several hypotheses have been proposed to explain why Seq-KD is effective:
\citet{gu2018non} point out that distilled data is less noisy and more deterministic than the original target data;
whereas \citet{zhang-etal-2018-analyzing} mention reduced fertility\footnote{The number of distinct target words aligned to a source word.} and distortion\footnote{Refers to word order, the number of crossed word alignments}.

More broadly, KD is useful in that it reduces the complexity of a dataset (a phenomenon known as `mode reduction,' \citealp{zhouunderstanding}).
As such, KD can be understood as a regularization technique: distilled training data for Seq-KD has lower entropy \citep{sen-etal-2023-self}, and consequently this greater regularity is easier for models to fit to.
In practice, this regularization acts by minimizing the target-token dependency in the output sequence --- i.e., how much context information from the target side is needed to generate a target token \citep{ren-etal-2020-study}.
This consequently leads to data that is more deterministic \citep{zhouunderstanding}
and to student models that exhibit less variation, so as to match these lower entropy distributions \citep{sen-etal-2023-self}. %
In that sense, Seq-KD can be argued to be similar to data diversification \citep{song2021data}: crucially its success as a regularization method can be imputed to the fact that KD does not restrict model capacity, unlike techniques such as dropout \citep{gordon2019explaining}.

While this can lead to higher quality translations, concerns have been raised that this very mechanism can also lead to poor translation performance on low-frequency words \citep{ding2021rejuvenating,ding2021understanding,DING2025101765} and reduced syntactic and lexical variation \citep{xu2021does}.
This regularization mechanism inherent to KD has other foreseeable detrimental effects.
As \citet{dankers-raunak-2025-memorization} discuss, by making the data more predictable, it stands to reason that students tend to memorize more than baseline models (as they overfit to frequent sequences) and tend to suffer from higher hallucination rates (as students can favor predictability on the target side over extrapolation of how a target token relates to the source, cf. \citealp{voita-etal-2021-analyzing}).
Similarly, although not specific to MT, recent work reports boosted benchmark scores under KD \cite{mansurov-etal-2025-data}, raising concerns about fair evaluation.
Another case concerns gender biases in MT models \citep{renduchintala-etal-2021-gender}: by favoring locally predictable patterns, distilled models tend to over-rely on gender stereotypes when translating occupations \citep{vamvas-sennrich-2021-contrastive}.
This last point is all the more concerning that there is broader evidence beyond MT that distillation amplifies biases without impact on the benchmark scores \citep{ahn-etal-2022-knowledge}.

In short, there is evidence that a large proportion of practitioners of KD in MT focus on KD as a tool for regularizing data rather than for compressing models. If this can lead to improved performances, there are also documented pitfalls --- higher hallucination and memorization rates, amplified biases and lower performances on rarer words --- that we should pay heed to.
Beyond the issues specific to regularization, there are other limitations concerning the usage of KD in the MT literature, in particular, regarding evaluation and coverage.

\subsection{Many roads not taken}
\label{sec:discussion_roads}

Many works reviewed here build on top of previous studies and, therefore, reuse the same datasets and metrics, keeping a rather narrow scope, which questions the generalization of the findings and leaves many questions unanswered.
We first focus on three aspects: datasets, evaluation metrics and language pairs.

\begin{figure}[h]
  \begin{subfigure}{\linewidth}
 \centering
 \includegraphics[width=0.9\linewidth]{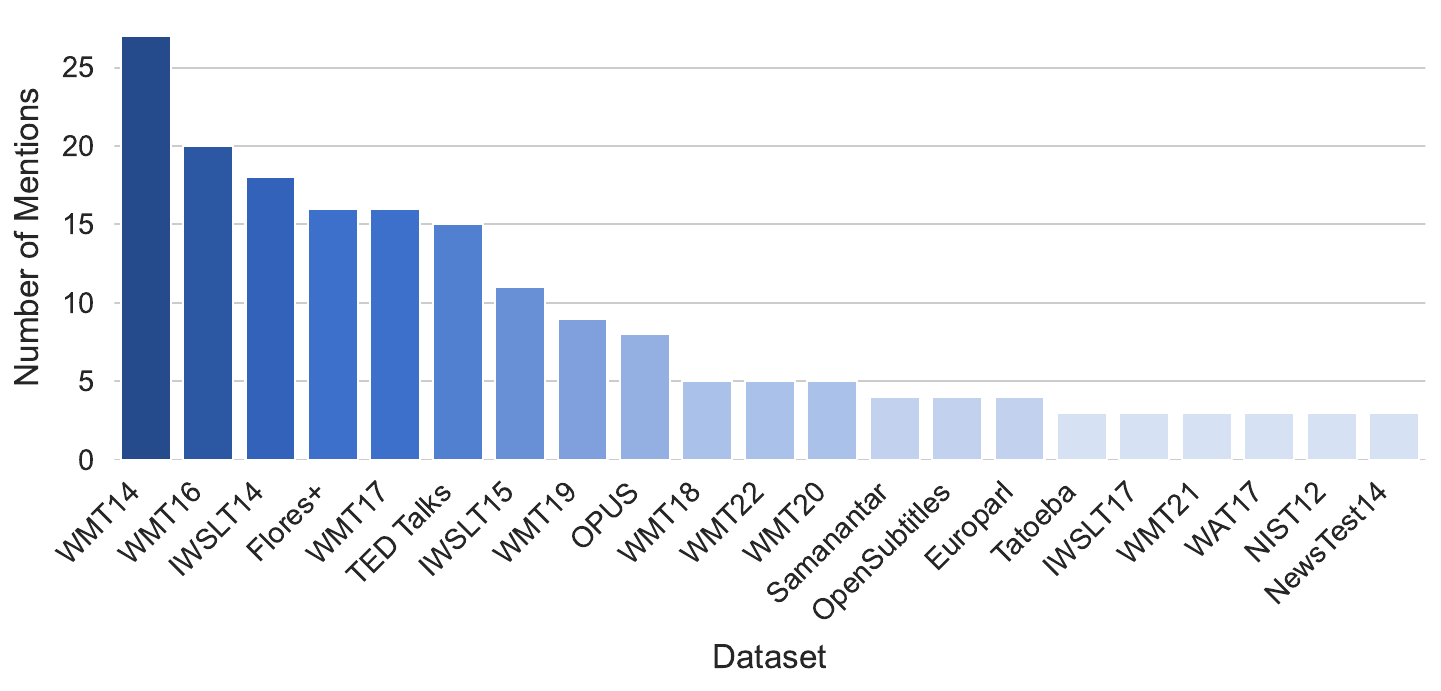}
 
     \caption{Dataset usage.}
     \label{fig:datasets}
\end{subfigure}

  \begin{subfigure}{\linewidth}
    \centering
    \includegraphics[width=0.7\linewidth]{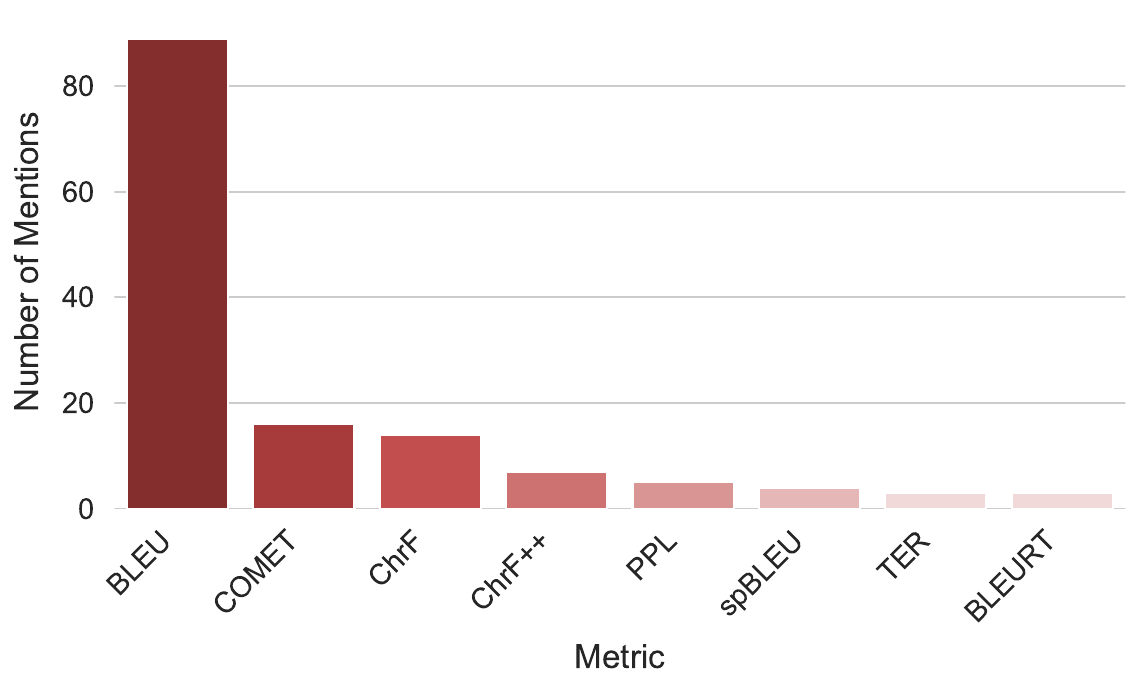}
    \caption{Metric usage.}
    \label{fig:metrics}
  \end{subfigure}
\caption{Frequency of dataset and metric usage in the surveyed papers. %
Only datasets and metrics that appear more than twice are shown.} %
  \label{fig:datasets-metrics}
\end{figure}

In \Cref{fig:datasets}, we list the \textbf{datasets} used across the 105 works surveyed and include a tally of how often they are encountered, ignoring datasets only mentioned once. 
As is apparent, while a handful of datasets are overwhelming more represented \cite[WMT 14, WMT 15, WMT 17, cf.][]{bojar-etal-2014-findings,bojar-etal-2015-findings,bojar-etal-2017-findings}, most of the datasets we cite are only listed in 2 or 3 publications.
Crucially, even the most frequently used dataset only covers a fourth of the works. We further examine why these datasets are so prevalent and find that several correspond to the same datasets and language pairs originally used by \citet{vaswani2017attention}.

\Cref{fig:metrics} counts the \textbf{evaluation metrics} mentioned in the surveyed papers. The majority of the works report BLEU scores \cite{papineni2002bleu}, either exclusively or complemented with other metrics. While more modern metric families such as COMET \cite{rei-etal-2020-comet} or ChrF \cite{popovic-2015-chrf,popovic-2017-chrf} should be preferred because they generally show higher correlation with human judgements than BLEU, the lack of a standard evaluation setup severely hinders the comparability of the proposed approaches. 
In short, despite the growing interest for KD in MT applications we have documented above, it is difficult to compare the relative improvement of any given technique given the lack of standard shared by the field.

\begin{figure*}[h]
 \centering
 \includegraphics[max width=\linewidth]{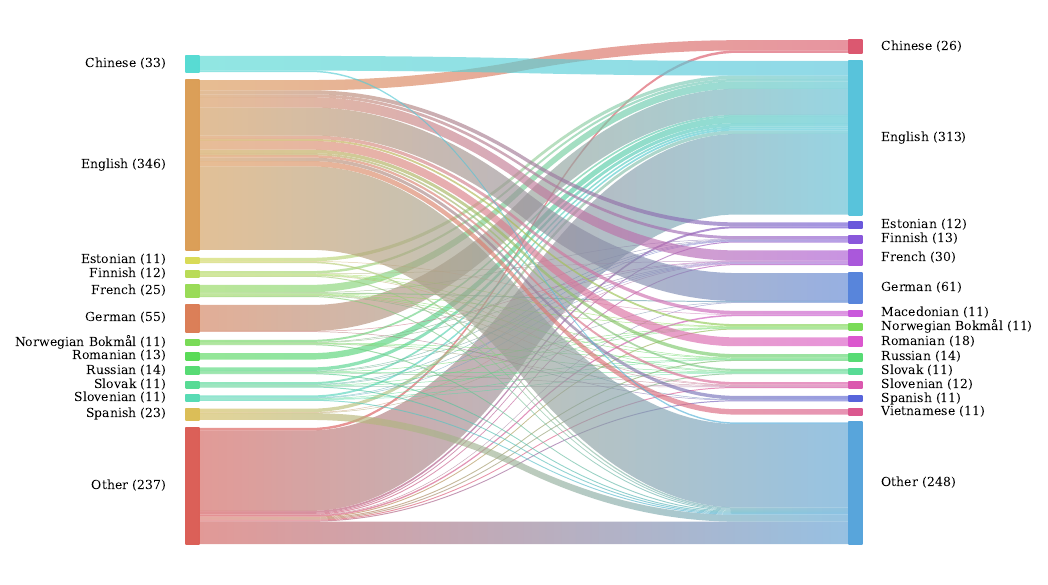}
 \caption{Frequency of translation directions considered in the surveyed papers, excluding papers focusing on massively MMT. The Sankey plot shows the distribution between source (on the left) and target (on the right) languages; source and target languages that appeared less than 10 times are grouped under `Other.'
 The width of a band indicates how often the corresponding language pair was attested. 
 Best viewed in color.
 }
 \label{fig:langpairs}
\end{figure*}

Additionally, we also remark that researchers elect to work on a large variety of \textbf{language pairs}, and that the field as a whole is still deeply committed to English-centric translation directions. 
\Cref{fig:langpairs} tallies all the language pairs mentioned in the surveyed body of work, besides those targeting massively multilingual translation.\footnote{This exclusion is motivated by practical reasons: while massively MMT applications correspond to only 4/105 works, they introduce over 1000 language pairs and as such are not good indicators of the general trends we observe.} 
A few key observations emerge from \Cref{fig:langpairs}: 
80.76\% of the works are dedicated to translation from English or into English --- in fact, translation from and into an non-English language corresponds to a minority of the language pairs mentioned in the works we reviewed.

On top of these issues that KD4MT inherits from MT, the field, being still quite nascent, has yet to be fully explored, and major research gaps remain, as hinted at in \Cref{fig:breakdown-heatmap}.
Although Word-KD and Seq-KD dominate the literature (92\% of surveyed works), there is no consensus on whether they outperform feature-based methods and under which conditions. Certain areas prominent in KD, such as adversarial KD \cite{chung2020feature,ZHANG2021107659} and graph-based KD \cite{he2019language,liu2023graph}, are also significantly underexplored for MT, despite their potential. Similarly, gradual learning approaches such as annealing KD \cite{jafari-etal-2021-annealing}, are limited in KD4MT. We hypothesize that this is because they face limitations similar to curriculum learning \cite{bengio2009curriculum}, where the key challenge lies in determining an effective ordering of training samples. In fact, curriculum learning often yields only marginal or inconsistent improvements compared to standard training, and in some cases may even impede convergence if not carefully designed \cite{surkov-etal-2022-data}. 

Likewise, few comprehensive studies exist to compare the different teacher--student architectures, and many applications remain underexplored.
A few exceptions exist: \citet{lai-etal-2021-lmu} apply \gls{selective-kd} \citep{wang-etal-2021-selective}, and \citet{yu-etal-2021-hw} use ensemble Seq-KD \citep{freitag2017ensemble} in multilingual settings. Beyond these cases, however, the applications of SOTA methods to specific applications remain rare.
We ought to expect that some time is needed before novel, cutting-edge KD methods (as outlined in \Cref{sec:algorithms}) are adopted for specific use-cases (\Cref{sec:apps}).

In summary, much remains to be done in order to get an adequately clear picture of what works and what doesn't in KD4MT.
On top of the uneven focus across application areas and algorithms, as introduced in \Cref{sec:discussion_tendencies}, the field has also not settled on common evaluation practices and much of what is published applies first and foremost to English-centric translation --- both issues that are already well-known in NLP as the cause of many pitfalls \citep{post-2018-call,10.1145/3442188.3445922,ducel-etal-2022-name}.
Not only does our survey outline clear research gaps that should be addressed, it also underscores the need for a clearer focus from the KD4MT community as to what are the best practices one should follow when it comes to evaluation.

\subsection{The role of LLMs}
 \label{sec:discussion_llms}

Finally, we focus on a specific research gap that has been attracting growing interest in the community, namely, the role of LLMs in KD4MT.
Currently, there is limited work on distilling LLMs specifically for MT. Rather than distilling from an LLM teacher, most recent studies on LLM-based MT train models of varying sizes from scratch; for example, TowerInstruct releases model variants of 7B and 13B parameters \cite{alvestower}.
Of the 105 surveyed papers, only 10 present LLM-based MT systems (9.5\%).
We hypothesize that this is due to several inherent challenges with LLMs, including scalability, computational cost, and lack of interpretability \cite{ganesh2025advances}. Moreover, most open-source LLMs still underperform task-specific MT models --- especially in the case of LRMT --- while the strongest systems %
are proprietary, limiting their use in non-black-box distillation settings, such as Word-KD.
Furthermore, traditional KD trains a smaller student with a similar architecture as its teacher, yet LLM research rarely builds such replicas at scale, for the same practical reasons.

Instead, we find in the literature that LLMs are used (i) as external knowledge sources (\Cref{sec:enriching}) and (ii) as synthetic data generators (\Cref{sec:lrmt}).
Their dual role enables richer and more flexible distillation paradigms. When data is scarce, LLMs extend coverage through Seq-KD, while in supervision-based approaches, they guide students with nuanced feedback via refinement and rationale generation.
Concretely for Seq-KD, practitioners first use an LLM to generate a distilled set and then proceed to incorporate it in two ways: (i) train a encoder-decoder model on that set (computationally cheaper) or (ii) fine-tune a smaller LLM (potentially better performance). To our knowledge, no study has systematically compared these two approaches.

Moving forward, we foresee two directions.
On the one hand, there is a growing shift towards smaller, more efficient LMs. More research is needed in order to transfer the efforts towards developing these models (e.g., TinyLlama, \citealp{zhang2024tinyllama}) to the specific context of MT, which is a promising direction for future work.
On the other hand, we anticipate a broader use of Seq-KD with  self-distillation (self-Seq-KD) for general-purpose LLMs: an LLM is used to generate diverse instruction sets and is then fine-tuned on its outputs \cite[e.g.][]{zhang2024enhancing}. %
Specifically, we expect that the boundary between self-Seq-KD and synthetic data generation will continue to blur, with synthetic datasets used not only for fine-tuning but in several stages of LLM training, such as pretraining \cite{kang2025demystifying}. 
Key open issues still remain, such as establishing how strong the teacher must be to yield useful data, and characterizing quality–quantity trade-offs in synthetic corpora.
A practical concern is however the over-reliance on synthetic corpora. To mitigate this, we advocate for clear documentation.
Moreover, %
as discussed in \Cref{sec:regularization}, Seq-KD's effectiveness  hinges on the fact that it simplifies the training signal, thereby reducing linguistic diversity and potentially amplifying existing biases. As the community shifts towards synthetic data, we should also be explicit about the implications this may have.

\section{Conclusion}
\label{sec:conclusion}
This survey situates KD4MT as more than a compression tool: it is a mechanism for shaping supervision, reducing data entropy, and reallocating capacity in ways that systematically alter generalization. Below we distill takeaways, articulating mechanisms, risks, and a concrete agenda for the next phase of work.

Across settings, Seq-KD acts as a mode reduction strategy: distilled targets are more deterministic and stabilize training, especially for NAT and SiMT, while freeing capacity within in the model. Word-KD remains a compelling alternative where compute or teacher access favors token-wise supervision, while feature-based KD (although comparatively underexplored in MT) offers a principled route to align internal representations.

For practitioners, the choice of method has to be driven by access constraints and resource budgets. Seq-KD is the most flexible when teachers are black-box or in architecture mismatch. When compute is tight or teacher internals are available, Word-KD with top-$n$ filtering is efficient. Deep-encoder/shallow-decoder students are generally adequate architectures that balance between performance and efficiency.

At the same time, given the clear research gaps we outlined in \Cref{sec:discussion_roads}, the field lacks systematic answers to several critical questions. While we can infer emerging best practices from what has been reported (cf. \Cref{sec:discussion_pracitcal}), we do not yet have a reliable mapping from teacher--student architecture choices (depth, width, tokenization and attention structure) to distillation efficacy, nor a consensus on when feature-based KD outperforms response-based methods. Safety and fairness remain insufficiently explored: the mode-reduction that improves learning can hinder diversity and amplify biases; multiple reports of increased hallucination and memorization suggest a need for further research. Finally, while LLMs are increasingly used as supervisors, rerankers, or data generators, we lack knowledge of quality--quantity trade-offs and contamination problems.

We argue that it is essential to focus on the diversity in the distilled set: too little entropy harms rare words.  This can be tested by exploring decoding techniques and sample diversity as well as evaluating rare-word recall against measured token entropy. Progress also depends on comparability. We recommend a minimal research protocol: evaluate on at least one high-resource and one truly low-resource pair, and always report metrics such as COMET\footnote{When applicable for the language pair, since neural metrics often perform poorly out of the box for low-resource languages \cite{dixit2023indicmt,falcao-etal-2024-comet}.} and ChrF, alongside BLEU. There is also a need to go beyond aggregate scores, by using rare-word accuracy measurements  \cite{DING2025101765}, bias probes \cite{vamvas-sennrich-2021-contrastive}, and hallucination stress tests \cite{guerreiro-etal-2023-hallucinations}. Distilled datasets or generation methods, sampling seeds, and student checkpoints should be released to enable reproduction, with explicit documentation of synthetic data sources.

The field’s next phase should prioritize careful experimental design. Looking ahead, we anticipate a substantially greater reliance on LLMs, which comes with a need for stronger controls against contamination, bias amplification, and evaluation drift.
Our accompanying database and glossary are intended to support that shift toward cumulative and insightful KD research for MT, to converge into a set of methods that can tap into KD's maximum potential.

\appendix
\appendixsection{Glossary of Key Terms}
\applabel{sec:glossary}

We list key terms along with definitions in the glossary below. We also provide page information, linking the first occurrence of a term in each subsection where it occurs.
 
\glsaddallunused
\printnoidxglossaries

\begin{acknowledgments}
This project has received funding from the European Union’s Horizon Europe programme (GA No 101070350) and from UK Research and Innovation (UKRI) under the UK government’s Horizon Europe funding guarantee (GA No 10052546).
This work is also supported by the Research Council of Finland through project No~353164. ``Green NLP -- controlling the carbon footprint in sustainable language technology.''
\end{acknowledgments}

\bibliographystyle{compling}
\bibliography{anthology_1,anthology_2,COLI_template}

\end{document}

%% file: glossary.tex
\newglossaryentry{adaptive-kd}{
  name=Adaptive KD,
  description={Word-level KD with multiple teachers where each teacher's weight is softmax-normalized from its negative perplexity on the data; introduced by \citet{saleh2020collective}}
}

\newglossaryentry{back-distillation}{
  name=Back-Distillation,
  description={After distilling a student, use that student to further improve the teacher; introduced by \citet{tan-etal-2019-multilingual}}
}

\newglossaryentry{capacity-curse}{
  name=Capacity Curse,
  description={A larger/stronger teacher does not necessarily yield a stronger student}
}

\newglossaryentry{capacity-gap}{
  name=Capacity Gap,
  description={Performance drop caused by size mismatch, where a smaller student lacks parameters to fit data as accurately as a larger teacher}
}

\newglossaryentry{complementary-kd}{
  name=Complementary KD,
  description={Multiple teachers trained on data subsets unseen by the student; the student distills via Word-KD and teachers are reset from the student each epoch; proposed by \citet{shao-feng-2022-overcoming}, used by \citet{roy-etal-2024-enhancing}}
}

\newglossaryentry{feature-kd}{
  name=Feature-based KD,
  description={Distillation from intermediate representations of the teacher (hidden states/feature maps)}
}

\newglossaryentry{multilinguality-curse}{
  name=Multilinguality Curse,
  description={Increasing the number of languages can hinder performance due to limited capacity \cite{conneau2020unsupervised}}
}

\newglossaryentry{mutual-distillation}{
  name=Mutual Distillation,
  description={Two or more models are trained simultaneously, each acting as teacher and student; \citet{huang2023towards}}
}

\newglossaryentry{response-kd}{
  name=Response-based KD,
  description={Distillation from the teacher's final predictions: token-level (Word-KD) or sentence-level (Seq-KD)}
}

\newglossaryentry{selective-kd}{
  name=Selective Distillation,
  description={Word-KD applied conditionally (e.g., only when the teacher outperforms the student); used by \citet{tan2019multilingual,wang-etal-2021-selective,zhang2023importance}}
}

\newglossaryentry{self-kd}{
  name=Self-Distillation,
  description={A model distills knowledge from itself, treating the student as the teacher}
}

\newglossaryentry{seq-interp}{
  name=Sequence-Level Interpolation,
  description={Seq-KD where the teacher proposes multiple candidates and the one closest to the reference is kept; \citet{kim-rush-2016-sequence}}
}

\newglossaryentry{seq-kd}{
  name=Sequence-Level KD,
  description={Distillation from the teacher's output translation to the student; \citet{kim-rush-2016-sequence}. Also called \emph{hard} or \emph{offline} distillation}
}

\newglossaryentry{topk-distillation}{
  name=Top-K Distillation,
  description={Distillation using only the teacher's top-$K$ probabilities instead of the full distribution; \citet{tan2019multilingual}}
}

\newglossaryentry{word-kd}{
  name=Word-Level KD,
  description={Distillation from the teacher's token probability distribution to the student; \citet{kim-rush-2016-sequence}. Also called \emph{soft} or \emph{online} distillation}
}

\newglossaryentry{annealing-kd}{
  name=Annealing Distillation,
  description={Distillation where the temperature of the teacher's output is progressively lowered \cite{jafari-etal-2021-annealing}}
}

\newglossaryentry{data-diversification}{
    name=Data Diversification,
    description={As introduced by \citet{NEURIPS2020_7221e5c8}, it consits of generating data via forward and back translation and training on the concatenation of the original and the newly created synthetic data.}
}